\documentclass[letterpaper, 10 pt, conference]{ieeeconf}  

\IEEEoverridecommandlockouts                              

\usepackage{amsmath}
\usepackage{amsfonts}
\usepackage{algorithmic}
\usepackage{array}
\usepackage{amssymb}
\usepackage{textcomp}
\usepackage{stfloats}
\usepackage{url}
\usepackage{verbatim}
\usepackage{graphicx}
\usepackage{cite}
\usepackage{booktabs}
\usepackage{upgreek}
\usepackage{pifont}

\usepackage{soul}
\usepackage{multirow}
\usepackage[table]{xcolor}
\PassOptionsToPackage{hyphens}{url}
\usepackage{hyperref}
\definecolor{darkred}{rgb}{0.8, 0.0, 0.0}
\definecolor{darkgreen}{rgb}{0.0, 0.5, 0.0} 

\usepackage[ruled]{algorithm2e}
\usepackage{dsfont}
\usepackage{xcolor}
\usepackage{subcaption}
\allowdisplaybreaks[4]

\usepackage{makecell}

\ifodd 1
\else

\fi

\title{\LARGE \bf
Learning from Distributed Eyes: Leveraging Collaborative Perception for Automated Model Adaptation
}

\author{Yanan Ma\textsuperscript{\rm 1}, Yihang Tao\textsuperscript{\rm 1}, Zhengru Fang\textsuperscript{\rm 1}, Zihan Fang\textsuperscript{\rm 1}, Yiqin Deng\textsuperscript{\rm 2} \\
    Xianhao Chen\textsuperscript{\rm 3}, and Yuguang Fang\textsuperscript{\rm 1}    
\thanks{The work was supported in part by the JC STEM Lab of Smart City funded by The Hong Kong Jockey Club Charities Trust under Contract 2023-0108, in part by the Research Grants Council of the Hong Kong SAR, China (Project No. CityU 11216324), and in part by the Hong Kong SAR Government under the Global STEM Professorship. 
The work of Y. Deng was supported in part by the National Natural Science Foundation of China under Grant No. 62301300 and in part by the Shandong Province Science Foundation under Grant No. ZR2023QF053.
The work of X. Chen was supported in part by the Research Grants Council of Hong Kong under Grant 27213824 and CRS HKU702/24.}
\thanks{$^{1}$Yanan Ma, Yihang Tao, Zhengru Fang, Zihan Fang, and Yuguang Fang are with the Hong Kong JC Lab of Smart City and the Department of Computer Science, City University of Hong Kong, Hong Kong, China. E-mail: yananma8-c@my.cityu.edu.hk, yihang.tommy@my.cityu.edu.hk, zhefang4-c@my.cityu.edu.hk, zihanfang3-c@my.cityu.edu.hk, my.fang@cityu.edu.hk.}
\thanks{$^{2}$Yiqin Deng is with the School of Data Science, Lingnan University, Tuen Mun, Hong Kong, China. E-mail: yiqindeng@ln.edu.hk.}
\thanks{$^{3}$Xianhao Chen is with the Department of Electrical and Electronic Engineering, University of Hong Kong, Hong Kong, China. E-mail: xchen@eee.hku.hk.}%
}

\begin{document}

\maketitle
\thispagestyle{empty}
\pagestyle{empty}

\begin{abstract}

In autonomous driving, perception models often struggle to generalize to new environments due to domain shifts. While unsupervised model adaptation offers a feasible solution without labor-intensive manual labeling, existing methods that rely solely on the ego-vehicle's data often lead to inferior pseudo-labeling performance. To address this critical issue, we propose \texttt{LDE}, Learning from Distributed ``Eyes", a novel framework that transforms collaborative perception (CP) into a source of high-quality supervision for model adaptation. This pseudo-labeling approach is hyperparameter-insensitive and relatively reliable, assuming CP often outperforms single-agent's perception. However, naively implementing this approach encounters (1) the communication bottleneck of sharing rich features under time and bandwidth constraints, (2) the view discrepancy between the CP view and the learner's Field of View (FoV), and (3) the unreliability even in CP-generated labels. To address these issues, we design an adaptation-oriented feature sharing mechanism that selectively transmits the most critical information for adaptation, an FoV filtering method that meticulously eliminates mismatched labels, and a curriculum learning strategy to progressively exploit pseudo labels. Extensive experiments on 3D object detection tasks demonstrate that \texttt{LDE} consistently outperforms both the pre-trained models and state-of-the-art unsupervised adaptation methods.

\end{abstract}

\section{INTRODUCTION}

\label{sec:intro}
While perception models in autonomous driving and robotics perform well in their training dataset~\cite{chen2024vehicle, hu2025collaborative}, they often struggle to generalize in the open real world~\cite{dosovitskiy2017carla, wang2020train, ma2026sense4fl}. This generalization gap necessitates continuous model adaptation to new environments. Nonetheless, traditional supervised model adaptation can be prohibitively expensive, requiring time-consuming and labor-intensive manual labeling, e.g., ground-truth bounding boxes and object classes in target scenes. Considering the case where models need to be adapted on an agent locally due to privacy and bandwidth constraints~\cite{fang2024r, fang2024pacp}, it is unrealistic to expect drivers or users, who are often not expert human annotators, to label data whenever they encounter new environments or changing conditions. Moreover, recent semi-automated annotation methods (e.g., leveraging powerful foundation models) may not be feasible either, as they typically require substantial computing/bandwidth resources, demand access to cloud APIs, and violate user privacy~\cite{ince2021semi, bai2025annotation, reza2025segbuilder}. This raises a fundamental question: \textit{Can we bypass manual labeling for perception model adaptation on autonomous agents?}

\begin{figure}[t!]
\begin{center}
\centering
\includegraphics[width=\columnwidth]{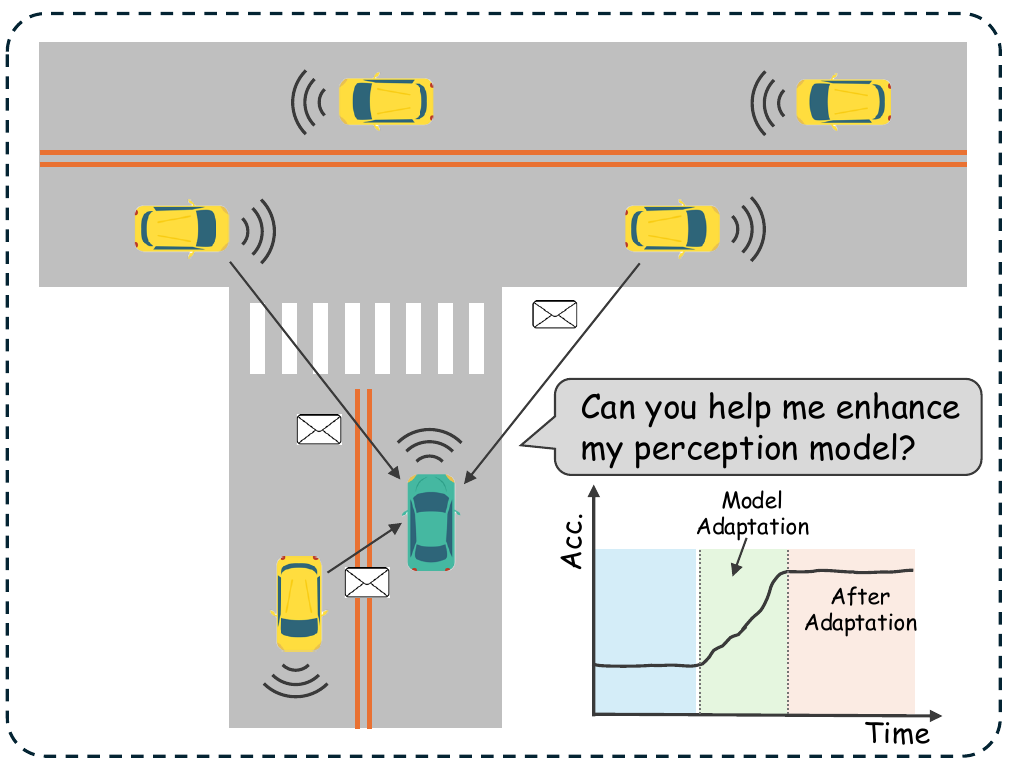}
\caption{Overview of leveraging CP for automated model adaptation. By leveraging features from surrounding CAVs, the ego vehicle generates high-quality pseudo-labels to improve its local perception model without manual annotation.}
\label{fig:system}
\end{center}
\end{figure}

To answer this question, unsupervised model adaptation or test-time adaptation (TTA) enables pre-trained models to be adapted to entirely unlabeled data. However, existing TTA solutions face significant shortcomings when applied to autonomous driving. \textit{First}, the quality of pseudo labels generated by TTA schemes can be unreliable as they are often highly sensitive to hyperparameter choices and exhibit significantly varying performance across different scenarios~\cite{ruan2024fully, cao2024exploring,boudiaf2022parameter}. \textit{Second}, the performance of these methods is fundamentally limited by the ego-centric views. Operating only on the ego vehicle's perspective makes it difficult to generate reliable labels for partially occluded objects or resolve perceptual ambiguities. This limitation of these methods, due to \textit{data quality}, inherently restricts their potential in pseudo-labeling.

Collaborative perception (CP) has emerged as a key technology for autonomous driving, which aggregates data from multiple vehicles to achieve perception accuracy far beyond that of a single agent~\cite{fang2024pacp, fang2024r, hu2025collaborative,tao2024directed, xu2022opv2v,hu2022where2comm,ma2026birdcast, ma2026update}. Vehicle-to-vehicle (V2V) systems for CP have been widely recognized in 5G systems and beyond (5G+)  to enhance road safety and traffic efficiency~\cite{chen2024vehicle, hu2024adaptive}. Crucially, we advocate that this high-fidelity output can naturally serve as ``teacher'' predictions to supervise the adaptation of an individual vehicle’s model. 
For instance, as illustrated in Fig.~\ref{fig:system}, under a limited communication budget, an ego connected and autonomous vehicle (CAV) can proactively leverage CP to generate high-quality pseudo-labels for model adaptation.
This pseudo-labeling approach is hyperparameter-insensitive and relatively reliable, assuming CP often outperforms a single vehicle’s perception. This simple yet effective strategy not only integrates seamlessly with 5G V2V systems but also enables reliable model adaptation with relatively accurate pseudo-labels derived from the ``combined wisdom'' of CAV.

However, existing CP schemes never leverage these CP prediction results for model adaptation purposes. Attempting to do so presents challenges regarding communication constraints, view discrepancies between the learner and CP results, and the reliability of pseudo-labeling. In response, we propose \texttt{LDE}, Learning from Distributed ``Eyes'', a fully automated model adaptation approach for 3D object detection with a multi-stage pipeline. To tackle the communication bottleneck, we introduce a selective, adaptation-oriented feature sharing mechanism. Instead of naively sharing all features, this approach reformulates feature selection as a multiple-choice knapsack problem (MCKP), prioritizing features that yield the highest utility score within the learner's FoV under the transmission budget. To resolve the learner-CP view discrepancy, we develop a FoV filtering method. It first conservatively shrinks bounding boxes to account for shape and range uncertainty, and then retains only the pseudo labels that are visible from the learner's egocentric perspective. Finally, to mitigate inherent pseudo-label noise, we employ a confidence-based curriculum learning strategy, allowing the model to adapt by initially focusing on the most reliable pseudo-labels before progressively processing others.


The main contributions of this paper are summarized as follows.
\begin{itemize}
    \item We propose a novel unsupervised model adaptation framework named \texttt{LDE}, which leverages CP as a source of ``teacher" supervision to generate high-quality pseudo labels and overcome the unreliability of single-agent adaptation.

    \item We design an adaptation-oriented feature-sharing approach by considering practical communication constraints, utilize FoV filtering to eliminate out-of-range labels caused by view discrepancies, and employ curriculum learning to enhance adaptation reliability.

    \item We conduct extensive performance evaluations on both simulated and real-world datasets, i.e., V2X-Sim and DAIR-V2X datasets, to demonstrate that our \texttt{LDE} framework outperforms non-adaptive baselines and existing unsupervised benchmarks on 3D object detection tasks.
\end{itemize}

\section{Related Work}
\subsection{Test-time Adaptation for Object Detection} 
Source-free unsupervised domain adaptation or TTA, which adapts models to unlabeled test data without requiring access to the source dataset~\cite{zhang2022memo, ruan2024fully}, is often demanded in autonomous driving and robotics, e.g., unsupervised 3D object detection tasks~\cite{yoo2025learning,ruan2024fully,you2022unsupervised,xia2025learning}. Yoo \textit{et al.}~\cite{yoo2025learning} developed an unsupervised adaptation scheme for 3D perception by learning from other predictions. However, this scheme can work only if others have better prediction quality than the learner. You \textit{et al.}~\cite{you2022unsupervised} exploited several repeated traversals of the same routes in the target domain to enhance unsupervised 3D object detection. Nonetheless, while a company can build a large-scale dataset through many repeated traversals, an individual vehicle (learner) may not be able to acquire such a local dataset. Xia \textit{et al.}~\cite{xia2025learning} devised DOtA, an automated pipeline for constructing object detection labels based on unlabeled CP datasets. However, this scheme exploits the shared pose and shape information of each CAV, which may not be accurately available and may not be effective for detecting objects other than CAVs, such as pedestrians and bicycles. In a nutshell, prior approaches, when applied to automated model adaptation in autonomous driving, have limited application scopes. More fundamentally, TTA methods often find it difficult to obtain reliable pseudo labels in the target domain~\cite{boudiaf2022parameter}. This motivates us to investigate how to utilize CP results as pseudo-labels for unsupervised model adaptation.

\subsection{Collaborative Perception}
Benefiting from the combined information from multiple CAVs, CP is widely recognized for its superiority over single-agent perception. A significant portion of existing CP research focuses on enhancing communication efficiency while preserving prediction quality~\cite{liu2020who2com,wang2020v2vnet,hu2022where2comm,tao2024directed, ma2026birdcast,ma2026update}. For instance, Who2com~\cite{liu2020who2com} employs a multi-stage handshake mechanism to compress information via matching scores. V2VNet~\cite{wang2020v2vnet} uses graph neural networks to aggregate information from nearby CAVs, while Where2comm~\cite{hu2022where2comm} utilizes the detection head to direct regions for sparse interactions. How2comm~\cite{yang2023how2comm} proposes to use a mutual information-aware mechanism for feature sparsification and a flow-guided strategy to compensate for temporal asynchrony. Similarly, PACP~\cite{fang2024pacp} develops a BEV-match mechanism to prioritize vehicles and optimize transmissions.
To address heterogeneity among agents, STAMP~\cite{gao2025stamp} introduces a scalable, task- and model-agnostic pipeline. It uses lightweight adapter-reverter pairs to transform BEV features between agent-specific models and a shared protocol domain, enabling efficient collaboration even when agents run different model architectures.
Despite these advancements in communication efficiency and model-agnostic fusion, none of these works employ CP for \textit{unsupervised adaptation of a vehicle's individual model}.

When directly applied to model adaptation, the existing CP methods lead to three issues: 1) the transmitted data may not necessarily be the most valuable results for model adaptation; 2) the out-of-range detection results not in the ego view of the learner can mislead model adaptation due to the view discrepancy; 3) learning may overfit toward unreliable pseudo labels, even when generated by CP.

\begin{figure*}[t!]
\begin{center}
\centering
\includegraphics[width=\textwidth]{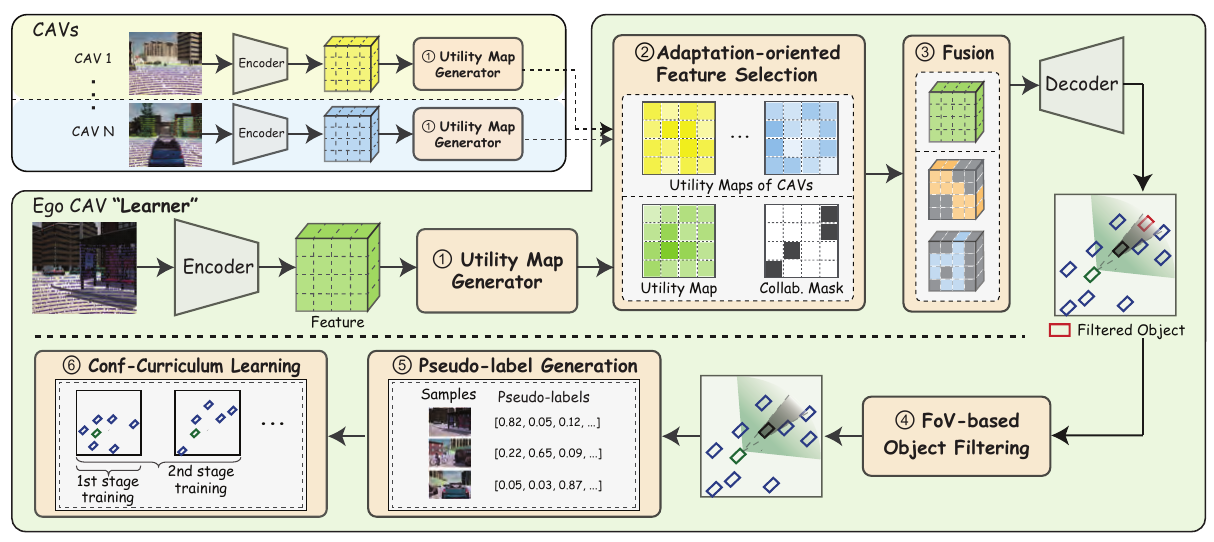}
\caption{Illustration of the \texttt{LDE} framework. The ego CAV (learner) collaborates with $N$ surrounding CAVs to generate high-quality pseudo labels for model adaptation. The framework utilizes adaptation-oriented feature selection and fusion based on spatial utility maps, followed by FoV-based pseudo label filtering to remove out-of-range detections. Finally, confidence-based curriculum learning progressively adapts the learner's perception model to mitigate the impact of corrupted labels.}
\label{fig:LDE_framework}
\end{center}
\end{figure*}

\section{Problem Definition}
\label{sec:method}

Given the perception model of an ego vehicle (the \textit{learner}) pre‑trained on a source dataset, our goal is to adapt it to an unlabeled target dataset $\mathcal{D}$. Formally, the objective is to find the ego vehicle's predictor $f^{\star}$ that minimizes the empirical risk on samples drawn from the target dataset 
\begin{equation}
f^{\star}
= \arg\min_{f\in\mathcal{F}} \frac{1}{\vert {\mathcal{D}}\vert }\sum_{({X}_{\rm{ego}},y_t)\in\mathcal{D}} \ell\bigl(y_t, f({X}_{\rm{ego}})\bigr),
\end{equation} 
where $({X}_{\rm{ego}}, y_t)$ denotes a sample from $\mathcal{D}$, $\vert {\mathcal{D}}\vert $ is the total number of samples in $\mathcal{D}$, $\mathcal{F}$ is a hypothesis class of predictors, and $\ell(\cdot,\cdot)$ is the detection loss.
However, in the target domain, the ground‑truth labels $y_t$ are unavailable, making it infeasible to optimize the above objective. 

To address the dilemma, we consider the case where the learner collaborates with $N$ surrounding CAVs in a vehicular network ($N$ and the set of CAVs can vary as the learner moves). Let ${X}_i$ denote the observations of the $i$-th CAV, which has many overlapped objects with ${X}_{\rm{ego}}$ in $\mathcal{D}$ from different perspectives and distances. By considering communication efficiency, each selected CAV transmits sparse intermediate feature $P_{i}$ extracted from ${X}_i$ to the learner.
Through CP, the learner can obtain the fusion results by
\begin{equation}
    \Tilde{y} = g \left({X}_{\rm{ego}},\left\{P_{i}\right\}_{i=1}^N\right),\label{mapping}
\end{equation}
where $g(\cdot)$ is a function that maps the ego vehicle's data and received features to a pseudo label. Since CP results are generally more accurate than those of the learner alone, $\Tilde{y}$ can be naturally used as pseudo labels to supervise the adaptation of the learner’s model.

\section{Proposed LDE Framework}
The proposed \texttt{LDE} framework, illustrated in Fig. \ref{fig:LDE_framework}, consists of five key steps: 1) Each CAV first extracts features with an encoder and then generates a \textit{spatial utility map}. 2) The learner selectively requests relevant features from neighboring CAVs under communication constraints. 3) The feature fusion module aggregates these features. 4) A FoV filtering method removes out-of-range pseudo labels that are not visible to the learner. 5) A confidence-based curriculum learning approach is used for progressive model adaptation. In what follows, we introduce the framework step by step.

\subsection{Feature and Utility Map Generation}
Each CAV extracts feature maps from its raw sensor data (e.g., 3D point clouds) using an encoder $\Phi_{\rm{Enc}}(\cdot)$, which can be represented in a bird's-eye view (BEV) and hence projected into a unified global coordinate system. For observation $X_i$ of the $i$-th CAV, the extracted feature map is $\mathbf{F}_i=\Phi_{\rm{Enc}}(X_i) \in \mathbb{R}^{H \times W \times C}$, where $H$, $W$, and $C$ denote the height, width, and number of channels, respectively. Similarly, given observation $X_{\rm{ego}}$, the feature map of the learner is given by $\mathbf{F}_{\rm{ego}}=\Phi_{\rm{Enc}}(X_{\rm{ego}}) \in \mathbb{R}^{H \times W \times C}$.

To select valuable features for sharing, each vehicle derives a spatial utility map for its feature map. For the $i$-th CAV, this is computed as $\mathbf{D}_i=\Phi_{\rm{Util}}(\mathbf{F}_i) \in [0,1]^{H \times W}$, where the utility generator $\Phi_{\rm{Util}}(\cdot)$ is implemented via a detection decoder. Similarly, the learner's utility map is $\mathbf{D}_{\rm{ego}}=\Phi_{\rm{Util}}(\mathbf{F}_{\rm{ego}}) \in [0,1]^{H \times W}$.

\subsection{Adaptation-oriented Feature Sharing and Fusion}
Given utility maps, we aim to maximize collaborative gain under a constrained communication budget. To achieve this, we first model wireless channels to determine data rates and then formulate feature selection as a budget-constrained optimization problem.

\noindent \textbf{Communication Modeling.} 
We consider an orthogonal frequency division multiple access (OFDMA) transmission scheme between surrounding CAVs and the learner. 
According to Shannon's channel capacity, the data rate $R_i$ between the $i$-th CAV and the learner is given by
\begin{equation}
    R_i = B\log_2\left(1+\frac{p_id_i^{-\alpha}|h_i|^2}{n_0}\right),
\end{equation}
where $B$ is the allocated bandwidth, $p_i$ is the transmit power of CAV $i$, $d_i$ is the distance between CAV $i$ and the learner, $\alpha$ is the path loss exponent, $|h_i|^2$ is the small-scale fading gain, and $n_0$ is the noise power.

\noindent \textbf{Feature Sharing Problem Formulation.}
Our objective is to maximize the total utility gain from feature sharing. Formally, the utility gap $[\mathbf{G}_{i}]_{j,k}$ is defined as:
\begin{equation}
    [\mathbf{G}_{i}]_{j,k} = \max\{[\mathbf{D}_{i}]_{j,k} -  [\mathbf{D}_{\rm{ego}}]_{j,k}, 0\},
\end{equation}
where $[\mathbf{D}_{i}]_{j,k}$ denotes the $(j,k)$-th element of the utility matrix $\mathbf{D}_{i}$ for the $i$-th CAV. A large utility gap characterizes a spatial grid perceptible to neighboring CAVs but uncertain for the learner.


To determine which features to transmit, we introduce a binary selection matrix $\mathbf{S}_i \in \{0,1\} ^{H \times W}$ for each CAV, where $[\mathbf{S}_i]_{j,k} = 1$ if the grid $(j,k)$ from CAV $i$ is selected, and $0$ otherwise. The corresponding feature sharing optimization is formulated as 
\begin{subequations}\label{p:selection}
    \begin{align} 
    \max_{\mathbf{S}_i} ~&\sum_{i=1}^N \sum_{j=1}^H \sum_{k=1}^W [\mathbf{S}_i]_{j,k} * [\mathbf{G}_{i}]_{j,k}\\
    \text { s.t. } ~  &\sum_{i=1}^N  [\mathbf{S}_i]_{j,k} \leq 1, ~\forall {j}, \forall{k},\label{cons:1}\\
    &\sum_{i=1}^N \frac{\sum_{j=1}^H \sum_{k=1}^W [\mathbf{S}_i]_{j,k} * \delta}{R_i} \leq T, \label{cons:latency} \\ 
    &[\mathbf{S}_i]_{j,k} \in \{0,1\}, ~\forall {i}, \forall {j}, \forall{k},\label{cons:binary}
    \end{align}
\end{subequations}
where $\delta$ denotes the data volume of a single feature grid, $T$ is the communication latency requirement. Constraint \eqref{cons:1} enforces that each feature grid is selected at most once, and Constraint \eqref{cons:latency} ensures that the total transmission latency does not exceed deadline $T$ (which is subject to vehicle contact time and spectrum resources in a vehicular network).

\noindent \textbf{Solution Approach.}
The feature-sharing problem we consider follows the structure of a multiple-choice knapsack problem (MCKP), which is NP-hard and highly challenging to solve \cite{kellerer2004introduction}. We resort to a heuristic algorithm to obtain the solution efficiently. First, to reduce computational complexity and enhance robustness, we partition each high‑resolution utility gap matrix $\mathbf{G}_i\in\mathbb{R}^{H\times W}$ into $C_h\times C_w$ non‑overlapping cells of size $a_h\times a_w$, where $C_h = {H}/{a_h}$ and $C_w = {W}/{a_w}$, with index set of cell $(p,q)$ being $\Omega_{p,q} =\{(j,k)|(p-1)a_h<j\le pa_h,(q-1)a_w<k\le qa_w\}$, 
for $p=1,\dots,C_h$ and $q=1,\dots,C_w$.
The cell-level utility gap at $(p,q)$ is then calculated by respecting the collaboration mask $\mathbf{S}$.

To facilitate solution finding, we define $\eta_{i,p,q} = {[\mathbf{G}_i^C]_{p,q}}/{\Delta t_i}$ as the utility-to-latency ratio, where $\Delta t_i = {\delta}/{R_i}$ is the transmission time for a single feature cell. We sort all candidate cells by $\eta_{i,p,q}$ in descending order and select them sequentially until either the latency budget $T$ is exhausted or all candidates have been evaluated. 
Upon completing the selection, the learner reconstructs the binary mask $\mathbf{S}_i$ for each CAV by 
\begin{equation}
    [\mathbf{S}_i]_{j,k} = 
\begin{cases}
1, & (j,k)\in\Omega_{p,q} ~\text{for the chosen cell}~(p,q),\\
0, & \text{otherwise.}
\end{cases}
\end{equation}
The learner then transmits $\mathbf{S}_i$ to the corresponding CAV. Finally, each selected CAV packs and transmits the resulting sparse feature map ${P}_i = \mathbf{S}_i \odot \mathbf{F}_i$ to the learner, where $\odot$ denotes the Hadamard product.

\noindent \textbf{Feature Fusion and Detection.}
Upon receiving the sparse features from collaborating CAVs, the learner applies a transformer-based fusion module that uses multi‑head attention to aggregate spatially aligned features. For notational simplicity, we denote the learner as index $i=0$ with $\mathbf{F}_0 = \mathbf{F}_{\rm ego}$.  The fused feature map is expressed as 
$\mathbf{F}_{\rm fuse}
= \mathrm{FFN}\Bigl(\sum_{i=0}^N \mathbf{W}_i \odot \mathbf{F}_i\Bigr)$,
where $\mathrm{FFN}(\cdot)$ is a feedforward network, and $\mathbf{W}_i$ is the attention weight map. 
Subsequently, feeding the fused feature map $\mathbf{F}_{\rm{fuse}}$ into the detection decoder yields a set of $M$ predictions, denoted as $\{(b_i, c_i)\}_{i=1}^M = \Phi_{\mathrm{Dec}}(\mathbf{F}_{\mathrm{fuse}})$. These predictions serve as the adaptation-oriented collaborative perception results. Each prediction tuple consists of a 3D bounding box $b_i$ and its associated confidence score $c_i \in [0, 1]$. The geometric configuration of each bounding box is parameterized as $b_i = (x_i, y_i, z_i, h_i, w_i, l_i, \theta_i)$. Finally, confidence-threshold filtering is employed to discard detections whose predicted score falls below the predefined threshold $\tau$ to eliminate low‐confidence outputs.


\subsection{FoV-based Pseudo Label Filtering}
Although the CP results can directly serve as pseudo-labels to supervise the learner's model adaptation, they may include objects outside the learner's effective FoV (e.g., in blind spots). Such out-of-view objects can mislead the adaptation process. To mitigate this view discrepancy, we introduce an FoV-based label filtering mechanism that retains only those detections truly visible to the learner. 

To enforce visibility consistency, we conduct a line-of-sight (LoS) analysis for LiDAR \cite{hagstrom2011line} on each candidate detection. This efficiently determines whether a detection lies along an unobstructed ray originating from the learner. Let $d_i$ denote the distance from the learner's sensor to the centroid of bounding box $b_i$. We first sort the set of bounding boxes $\{b_i\}$ in ascending order of $d_i$ to prioritize closer objects. Next, because cuboidal bounding boxes often overshoot the boundaries of real-world objects with curved surfaces, we conservatively shrink each box $b_i$ by a scaling factor $\epsilon_i \in (0,1]$. Specifically, we scale the spatial dimensions of the original bounding box to generate a shrunken counterpart $\hat{b}_i = (x_i, y_i, z_i, \epsilon_i h_i, \epsilon_i w_i, \epsilon_i l_i, \theta_i)$. To account for growing localization uncertainty at longer ranges, we define this distance-based ratio as
\begin{equation}
    \epsilon_i
= \epsilon_{\min} + (\epsilon_{\max}-\epsilon_{\min})\,e^{-\lambda\,d_i}.
\end{equation}
Consequently, more distant boxes (i.e., larger $d_i$) are shrunk more aggressively ($\epsilon_i \to \epsilon_{\min}$), thereby reducing false positives caused by noisy long-range detections. Ultimately, this conservative shrinkage mitigates potential occlusion errors arising from bounding-box inaccuracies and shape mismatches.

Next, we check \textit{visibility} by ray‐casting from the learner's sensor origin toward the centroid and vertices of $\hat{b}_i$. 
As long as at least one ray reaches $\hat{b}_i$ unobstructed by closer objects, this bounding box is deemed (partially) visible to the learner. Finally, we verify \textit{non-occlusion} by ensuring that $\hat{b}_i$ does not overlap with any previously accepted box $\hat{b}_j$ with $d_j<d_i$ in the BEV projection, which otherwise may indicate misdetections. By retaining only those $b_i$ that satisfy both visibility and non-occlusion, we obtain a filtered pseudo-label set aligned with the learner's ego-view, thereby facilitating reliable model adaptation.

\begin{table*}[!t]
\centering
\setlength{\tabcolsep}{3mm}  
\caption{\textbf{Comparative results under different pre-trained model accuracy on V2X-Sim and DAIR-V2X datasets.} We report test accuracies AP@0.3 and AP@0.5 after 10 epochs with best in \textbf{bold}, second-best \underline{underlined}, except for the fully-supervised (upper bound).  $\downarrow$ represents performance degradation compared with the pretrained (non-adaptive) model. Setup 1 and Setup 2 represent adaptation cases from pre-trained models with different initial accuracies, respectively.}
\label{table:baseline}
\resizebox{\textwidth}{!}{
\renewcommand{\arraystretch}{1}
\begin{tabular}{lcccccccc}
\toprule
& \multicolumn{4}{c}{\textbf{V2X-Sim}} & \multicolumn{4}{c}{\textbf{DAIR-V2X}} \\
\cmidrule(lr){2-5}\cmidrule(lr){6-9}
 \textbf{Method} & \multicolumn{2}{c}{\textbf{Setup 1}} & \multicolumn{2}{c}{\textbf{Setup 2}} & \multicolumn{2}{c}{\textbf{Setup 1}} & \multicolumn{2}{c}{\textbf{Setup 2}} \\
\cmidrule(lr){2-3}\cmidrule(lr){4-5}\cmidrule(lr){6-7}\cmidrule(lr){8-9}
& \multicolumn{1}{c}{AP@0.3} & \multicolumn{1}{c}{AP@0.5} & \multicolumn{1}{c}{AP@0.3} & \multicolumn{1}{c}{AP@0.5} & \multicolumn{1}{c}{AP@0.3} & \multicolumn{1}{c}{AP@0.5} & \multicolumn{1}{c}{AP@0.3} & \multicolumn{1}{c}{AP@0.5}\\

 Pretrained & 55.69 &49.06& 68.79 &60.31 & 46.01  & 41.22 &  55.61  & 50.37\\
 \midrule
 AdaBN & ~~48.12$\downarrow$ & ~~41.88$\downarrow$ & ~~63.22$\downarrow$ & ~~54.73$\downarrow$ & ~~38.12$\downarrow$ & ~~33.29$\downarrow$ & ~~48.64$\downarrow$ & ~~44.08$\downarrow$\\
 ST & ~~51.66$\downarrow$ & ~~45.79$\downarrow$ & ~~67.71$\downarrow$ & ~~58.43$\downarrow$ & ~~40.81$\downarrow$ & ~~35.26$\downarrow$ & ~~50.12$\downarrow$ & ~~46.77$\downarrow$ \\
 SN & 56.42 & 51.00 & 70.51 & 61.53& ~~42.33$\downarrow$ & ~~37.29$\downarrow$ & 57.38 & 52.14\\
 CPD & 57.71 & 52.84 & 71.85 & 63.89 & 48.79 & 44.10 & 58.02 & 54.71 \\
 DOtA  &59.17 & \underline{55.37} & 74.18 & 68.01 & 50.82 & 45.79 & 62.67 & 58.84 \\
 CP4Adaptation &~~44.38$\downarrow$ &~~37.73$\downarrow$ &~~59.36$\downarrow$ &~~51.19$\downarrow$ &~~39.78$\downarrow$ &~~36.70$\downarrow$ &~~48.81$\downarrow$ 
 &~~44.33$\downarrow$\\
 \textbf{LDE (Ours)} & \underline{63.33} & 55.24 & \underline{77.49} & \underline{71.27} & \underline{55.60} & \underline{49.11} & \underline{67.28} & \underline{63.41} \\
\textbf{LDE-Full (Ours)} & \textbf{69.05} & \textbf{61.72} & \textbf{83.93} & \textbf{75.73} &\textbf{59.01} &\textbf{52.25} & \textbf{69.00} &\textbf{64.98}\\
 Upper Bound & 86.17 & 81.79  & 86.93 & 82.34 & 70.23 & 65.76& 71.21 & 66.29 \\
\bottomrule
\end{tabular}
}
\end{table*}

\subsection{Confidence-based Curriculum Learning}

Upon model adaptation, we introduce a curriculum learning strategy to adapt the model in an ``easy-to-hard'' manner \cite{bengio2009curriculum, zhu2023curricular}. Rather than relying on a single, fixed confidence threshold, we employ a dynamic threshold $\tau_c$. Specifically, we initialize the adaptation process with a high confidence threshold to construct a high-quality subset of pseudo-labels, comprising only the most certain detections. Subsequently, we gradually decay this threshold to introduce more lower-confidence samples. This enables the model to smoothly propagate knowledge learned from the highly reliable initial set to broader, more challenging data distributions. Ultimately, this curriculum-based approach prevents early-stage model corruption and ensures a more stable and robust adaptation process.

\section{Experiments}

\subsection{Experimental Setup}

\noindent \textbf{Datasets.}
We conduct experiments on both simulated and real-world datasets, i.e., V2X-Sim dataset \cite{li2022v2x} and DAIR-V2X dataset \cite{yu2022dair}.

\begin{itemize}
    \item \textbf{V2X-Sim dataset.} V2X-Sim \cite{li2022v2x} is a simulated V2X collaborative perception dataset simulated using SUMO and CARLA~\cite{dosovitskiy2017carla}. It comprises 10,000 frames of 3D LiDAR point clouds captured from 5 CAVs, alongside 501,000 annotated 3D bounding boxes. We discretize the 3D point clouds into a BEV map with a size of $(256, 256, 13)$, and the resolution is 0.4 m/pixel in both length and width.

    \item \textbf{DAIR-V2X dataset.} DAIR-V2X \cite{yu2022dair} is a real-world collaborative perception dataset wherein each sample captures synchronized data from a vehicle and an infrastructure node. The effective perception range spans $201.6 \text{ m} \times 80 \text{ m}$. We represent the BEV map with size of $(200, 504, 64)$ and the resolution is 0.4 m/pixel.


\end{itemize}

\begin{table*}[h]
\centering
\setlength{\tabcolsep}{2.3mm}  
\caption{\textbf{Comparison of label quality on V2X-Sim and DAIR-V2X datasets.} We report the Recall and Precision of IoU@0.3 and IoU@0.5, with the best results shown in \textbf{bold}, second-best \underline{underlined}.}
\label{tab:label_quality}
\resizebox{\textwidth}{!}{
\renewcommand{\arraystretch}{1}
\begin{tabular}{lcccccccc}
\toprule
& \multicolumn{4}{c}{\textbf{V2X-Sim}} & \multicolumn{4}{c}{\textbf{DAIR-V2X}} \\
\cmidrule(lr){2-5}\cmidrule(lr){6-9}
 \textbf{Method} & \multicolumn{2}{c}{\textbf{Recall}} & \multicolumn{2}{c}{\textbf{Precision}} & \multicolumn{2}{c}{\textbf{Recall}} & \multicolumn{2}{c}{\textbf{Precision}} \\
\cmidrule(lr){2-3}\cmidrule(lr){4-5}\cmidrule(lr){6-7}\cmidrule(lr){8-9}

&  IoU@0.3 &IoU@0.5 &  IoU@0.3 & IoU@0.5 &  IoU@0.3 &IoU@0.5 &  IoU@0.3 & IoU@0.5\\

Pretrained  & 69.09& 66.37 &59.57 & 53.78 & 65.30 & 60.29 & 47.81 & 44.73\\
 \midrule
AdaBN & ~~59.28$\downarrow$ & ~~52.82$\downarrow$ & ~~43.79$\downarrow$& ~~38.53$\downarrow$ & ~~58.62$\downarrow$ & ~~52.50$\downarrow$ &~~39.56$\downarrow$ &~~36.18$\downarrow$ \\
ST & ~~67.39$\downarrow$ & ~~63.02$\downarrow$ & ~~42.39$\downarrow$ & ~~37.95$\downarrow$ & ~~63.42$\downarrow$ & ~~57.10$\downarrow$ & ~~36.56$\downarrow$ & ~~34.86$\downarrow$ \\
SN & 74.88 & ~~64.69$\downarrow$ & ~~52.62$\downarrow$ & ~~47.21$\downarrow$ & 68.65 & 60.45 & ~~43.85$\downarrow$ & ~~41.05$\downarrow$\\
CPD  & 77.03 & 68.12 & ~~57.89$\downarrow$ & 55.71 & 69.90  & 61.17  & 49.16 & 45.81\\
DOtA & 78.51 & 69.62 & 62.43 & 58.78 & 70.80  & 61.97  & 50.56  & 47.71\\
CP4Adaptation & 77.81 & 68.67 & ~~40.41$\downarrow$ & ~~36.09$\downarrow$ & {69.53}  & {60.22}  & ~~37.15$\downarrow$ &~~34.82$\downarrow$ \\
\textbf{LDE (Ours)} &\underline{79.20} & \underline{71.79}& \underline{64.72} & \underline{60.78} & \underline{71.20}  & \underline{63.01}  & \underline{53.48}  & \underline{50.65} \\
\textbf{LDE-Full (Ours)} &  \textbf{87.40} & \textbf{85.01} & \textbf{79.30} & \textbf{76.23} & \textbf{75.90} & \textbf{69.63} & \textbf{61.88} & \textbf{57.31}\\
\bottomrule
\end{tabular}
}
\end{table*}

\noindent \textbf{Implementation Details.}
For a LiDAR-based 3D object detection task, we conduct experiments with PointPillars \cite{lang2019pointpillars} as the default detector.
The bandwidth is $B = 20$ MHz and the transmit power is $5$ W.
The path loss exponent $\alpha$ is set to 2.3,  with a noise power of $n_0 = -120$ dBm. 
The time requirement is $T = 300$ ms. 
The data is split into model adaptation, validation, and test sets, with a ratio of 8:1:1. 
First, we pre-train the detector on the OPV2V dataset \cite{xu2022opv2v} to a predefined accuracy. Then, we perform inference on the adaptation dataset to obtain CP results by sharing the selected features with the learner under the communication constraint, which serve as pseudo-labels. 
Label filtering is conducted with $\epsilon_{\max} = 0.95$, $\epsilon_{\min} = 0.8$, and $\lambda=1$.
For our confidence-based curriculum learning, we adapt the model over 10 epochs. We start with a high confidence threshold of $\tau_c = 0.75$ for the first 3 epochs, then relax it to $\tau_c = 0.5$ for epochs 4-7, and finally use $\tau_c = 0.25$ for the remaining epochs.
We adapt the model using Adam optimizer with lr=$1e^{-5}$. All experiments are conducted on a server with 2 Intel(R) Xeon(R) Silver 4410Y CPUs, 4 NVIDIA RTX A5000 GPUs, and 512 GB RAM.
The performance of all methods is evaluated using average precision (AP) at 0.3 IoU (AP@0.3) and 0.5 IoU (AP@0.5) for detection accuracy.

\subsection{Quantitative Evaluation}

\noindent \textbf{Baselines.}
We evaluate our \texttt{LDE} framework against other model adaptation baselines: \textbf{AdaBN} \cite{li2016revisiting}, \textbf{ST} (self-training)~\cite{roychowdhury2019automatic}, \textbf{SN} (statistical normalization) \cite{wang2020train}, \textbf{CPD} \cite{wu2024commonsense}, \textbf{DOtA} \cite{xia2025learning}, and a fully supervised \textbf{Upper Bound}. 
For a fair comparison, all methods use PointPillars as the backbone. 
We also include two additional baselines: \textbf{CP4Adaptation} employs the CP results directly from the Where2comm framework \cite{hu2022where2comm} as pseudo labels, and \textbf{LDE-Full}, which is an implementation of \textbf{LDE} with an unlimited communication budget.



\noindent \textbf{Baseline Comparison.} 
The quantitative evaluations are detailed in Table \ref{table:baseline}, which reports the adaptation performance across two distinct initial model accuracies (Setup 1 and Setup 2). We observe that existing TTA baselines exhibit a critical weakness: they exploit knowledge only from the ego-vehicle's captured data. Consequently, methods such as AdaBN, ST, and SN fail to adapt effectively to the open-world setting, often resulting in significant performance degradation ($\downarrow$) compared to the non-adaptive pretrained model. 
While more advanced methods, such as CPD and DOtA, achieve modest performance gains, our proposed \texttt{LDE} framework consistently and substantially outperforms them across all setups on both the V2X-Sim and DAIR-V2X datasets. Specifically, on the real-world DAIR-V2X dataset under Setup 1, \texttt{LDE} achieves an AP@0.3 of 55.60\% and an AP@0.5 of 49.11\%, demonstrating robust real-world generalization. Furthermore, compared with CP4Adaptation, our approach yields striking absolute performance gains—such as an 18.95\% increase in AP@0.3 and a 17.51\% increase in AP@0.5 under Setup 1 on the V2X-Sim dataset. This demonstrates that naively implementing existing collaborative perception schemes for model adaptation is highly ineffective. Finally, our unconstrained \texttt{LDE-Full} variant consistently achieves the highest unsupervised accuracy, significantly narrowing the performance gap to the fully supervised Upper Bound.


\noindent \textbf{Label Quality Analysis.} 
To evaluate the quality of the generated pseudo-labels, we analyze the recall and precision across various approaches on both the V2X-Sim and DAIR-V2X datasets. As detailed in Table \ref{tab:label_quality}, \texttt{LDE} consistently yields superior pseudo-labels compared to existing unsupervised methods. Notably, \texttt{LDE} maintains robust performance and achieves a high recall of 79.20\% and a precision of 64.72\% on the V2X-Sim dataset at IoU@0.3. This substantial enhancement is primarily attributable to the multi-vehicle collaborative architecture of \texttt{LDE}: by aggregating spatially diverse features, the framework captures richer, multi-view object representations, thereby mitigating occlusions and reducing false negatives. 


\begin{figure*}[t]
\centering
\includegraphics[width=0.98\textwidth]{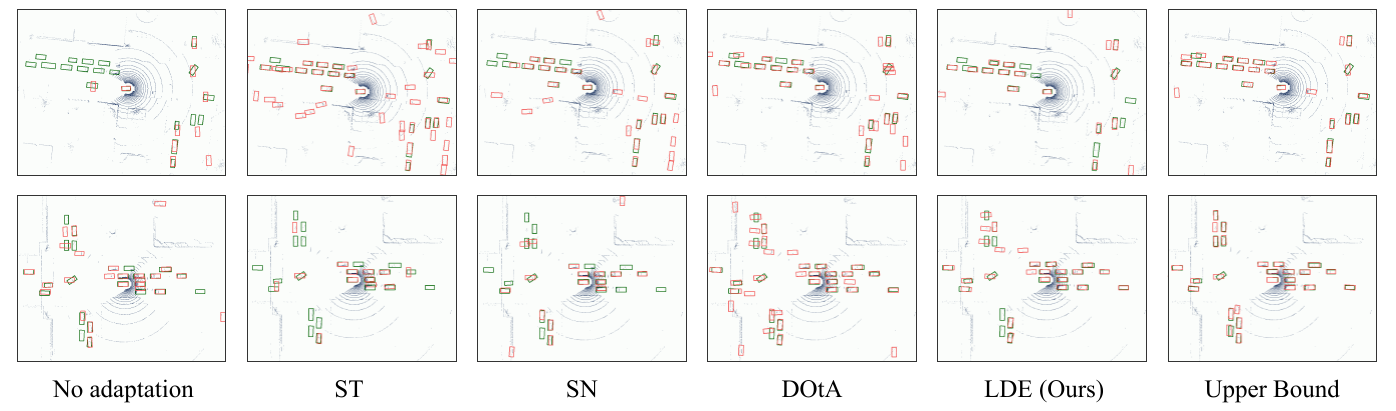}
\caption{Visualization of 3D detection results on the V2X-Sim dataset, comparing the performance of the adapted model with that of the original model and some baselines. \textcolor{darkred}{Red} bounding boxes denote the model predictions, while \textcolor{darkgreen}{green} ones represent the ground truth (GT).}
\label{fig:vis}
\end{figure*}

\begin{table}[t]
    \centering
    \caption{Ablation study on key components of \texttt{LDE} on V2X-Sim dataset. Best in \textbf{bold}, second-best \underline{underlined}.}
    \label{tab:components}
    \resizebox{1\columnwidth}{!}{
    \renewcommand{\arraystretch}{1}
    \begin{tabular}{ccccc}
        \toprule
        \multirow{2}{*}{\textbf{Adapt FS}} & \multirow{2}{*}{\textbf{FoV-Filter}} & \multirow{2}{*}{\textbf{CCL}} & \multicolumn{2}{c}{\textbf{V2X-Sim}} \\ 
        \cmidrule(lr){4-5}
         & & & AP@0.3 & AP@0.5 \\
         & & & 68.79 & 60.31 \\
        \midrule
        \checkmark & \checkmark & \texttimes & \underline{75.92} & \underline{69.75} \\
        \checkmark & \texttimes & \checkmark  & ~~63.86$\downarrow$ & ~~55.45$\downarrow$ \\
        \texttimes & \checkmark & \checkmark & 74.37 & 67.78\\
        \midrule
        \checkmark & \checkmark & \checkmark  & \textbf{77.49} & \textbf{71.27} \\
        \bottomrule
    \end{tabular} 
    }
\end{table}

\subsection{Ablation Study}

\noindent \textbf{Effect of Key Modules.} 
We first analyze the impact of our three core components, as shown in Table~\ref{tab:components}. Replacing our adaptation-oriented feature sharing (\textbf{Adapt FS}) with a random selection strategy degrades performance, as this wastes bandwidth on irrelevant data and lowers pseudo-label quality. Similarly, removing the FoV-based pseudo-label filtering (\textbf{FoV-Filter}) is highly detrimental, as it introduces a severe perceptual mismatch by forcing the model to learn from misleading labels outside its FoV. Finally, removing the confidence-based curriculum learning (\textbf{CCL}) in favor of a single-shot adaptation also harms performance, confirming our curriculum learning is essential for stable adaptation by allowing the model to learn from the most reliable labels first and preventing early-stage error accumulation.

\begin{table}[t]
\centering
\caption{Effects of the different communication constraints on V2X-Sim and DAIR-V2X datasets. Best in \textbf{bold}, second-best \underline{underlined}.}
\label{tab:comm}
\resizebox{1\columnwidth}{!}{
\renewcommand{\arraystretch}{1}
\begin{tabular}{lcccc}
\toprule
\makecell{\textbf{Comm.}\\\textbf{Budget}} & \multicolumn{2}{c}{\textbf{V2X-Sim}} & \multicolumn{2}{c}{\textbf{DAIR-V2X}} \\
\cmidrule(lr){2-3}\cmidrule(lr){4-5}

&  AP@0.3 &AP@0.5 &  AP@0.3 & AP@0.5 \\
Pretrained  & 68.79 & 60.31 & 55.61 & 50.37\\
 \midrule
$T$ = 100 ms & 73.01 & 64.44 & 60.41 & 55.27 \\
$T$ = 300 ms & \underline{77.49} & \underline{71.27} & \underline{67.28} & \underline{63.41} \\
$T$ = 500 ms & \textbf{78.92} & \textbf{72.87} & \textbf{68.92} & \textbf{64.33} \\
\bottomrule
\end{tabular}
}
\end{table}

\noindent \textbf{Effect of Communication Budget.} 
We further investigate the impact of the communication time budget on the \texttt{LDE} framework, with results summarized in Table \ref{tab:comm}. As time allocation increases, the learner is able to aggregate richer collaborative information and improve the quality of generated pseudo-labels, thereby leading to superior model adaptation performance.

\noindent \textbf{Effect of Parameters for Label Filtering.}
Table~\ref{tab:confidence} presents ablation results for label-filtering parameters on model adaptation. Adjusting the bounding-box shrinkage factor has little effect on overall detection accuracy. By contrast, the confidence threshold exhibits a clear trade-off: setting it too low admits noisy pseudo labels and degrades adaptation, while setting it too high excludes informative labels and likewise reduces performance.

\subsection{Qualitative Analysis}


\noindent \textbf{Visualization of Detection Results.}
To qualitatively demonstrate the effectiveness of our approach, Fig. \ref{fig:vis} visualizes the 3D detection results of models adapted by our \texttt{LDE} framework alongside several competitive benchmarks. As observed, existing baselines frequently struggle in complex driving scenes, producing numerous false positives and suffering from poor bounding box localization. In contrast, our \texttt{LDE} framework successfully leverages distributed spatial features to significantly suppress erroneous detections, delivering more accurate and robust detection results.


\begin{table}[t]
\centering
\caption{Effects of the different parameters for label filtering on the V2X-Sim dataset. Best in \textbf{bold}, second-best \underline{underlined}.}
\label{tab:confidence}
\resizebox{\columnwidth}{!}{
\renewcommand{\arraystretch}{1}
\begin{tabular}{lcccc}
\toprule
& \multicolumn{4}{c}{\textbf{V2X-Sim}}\\
\cmidrule(lr){2-5}
\textbf{Conf.} & \multicolumn{2}{c}{$\epsilon_{\mathrm{max}} =0.95$, $\epsilon_{\mathrm{min}} = 0.8$} & \multicolumn{2}{c}{$\epsilon_{\mathrm{max}} = 1$, $\epsilon_{\mathrm{min}} = 1$} \\
\cmidrule(lr){2-3}\cmidrule(lr){4-5}

&  AP@0.3 &AP@0.5 &  AP@0.3 & AP@0.5 \\

Pretrained  & 68.79 & 60.31 & 68.11 & 59.86 \\
 \midrule
$\tau$ = 0.15 & 76.23 &  70.04 & 75.45 & 69.75 \\
$\tau$ = 0.25 & \textbf{77.49} & \textbf{71.27} & \textbf{77.00} & \textbf{70.78} \\
$\tau$ = 0.35 & \underline{77.02} & \underline{70.63} & \underline{76.34} & \underline{70.27} \\
\bottomrule
\end{tabular}
}
\end{table}

\section{Conclusion}

In this paper, we have addressed the challenge of adapting perception models to new environments without the need for manual labeling for autonomous driving. We have introduced the Learning from Distributed “Eyes” (\texttt{LDE}) framework, which leverages collaborative perception to generate high-quality pseudo labels, enabling automated model adaptation. First, we have designed an adaptation-oriented feature-sharing mechanism to produce accurate predictions under communication constraints. Then, we have developed a field-of-view filtering scheme to mitigate view discrepancies among vehicles. Finally, we have introduced a confidence-based curriculum learning strategy to stabilize the adaptation process by managing inherent label noise. Extensive simulations have demonstrated that the proposed \texttt{LDE} framework delivers robust, consistent performance improvements over pre-trained models and existing unsupervised adaptation methods, paving the way for reliable perception in real-world autonomous driving scenarios.

While we choose 3D object detection in autonomous driving as the subject of study, our approach has the potential to benefit a wide range of vision tasks, such as depth estimation and segmentation, as well as other applications, such as drone and robotic systems.










\bibliographystyle{IEEEtran}
\bibliography{reference}

@inproceedings{hagstrom2011line,
  title={Line-of-sight analysis using voxelized discrete lidar},
  author={Hagstrom, Shea and Messinger, David},
  booktitle={Laser Radar Technology and Applications XVI},
  volume={8037},
  pages={104--114},
  year={2011},
  organization={SPIE}
}

@article{li2022v2x,
  title={{V2X-Sim}: Multi-agent collaborative perception dataset and benchmark for autonomous driving},
  author={Li, Yiming and Ma, Dekun and An, Ziyan and Wang, Zixun and Zhong, Yiqi and Chen, Siheng and Feng, Chen},
  journal={IEEE Robotics and Automation Letters},
  volume={7},
  number={4},
  pages={10914--10921},
  year={2022},
  publisher={IEEE}
}

@inproceedings{yu2022dair,
  title={{DAIR-V2X}: A large-scale dataset for vehicle-infrastructure cooperative {3D} object detection},
  author={Yu, Haibao and Luo, Yizhen and Shu, Mao and Huo, Yiyi and Yang, Zebang and Shi, Yifeng and Guo, Zhenglong and Li, Hanyu and Hu, Xing and Yuan, Jirui and others},
  booktitle={Proceedings of the IEEE/CVF Conference on Computer Vision and Pattern Recognition},
  pages={21361--21370},
  year={2022}
}

@inproceedings{dosovitskiy2017carla,
  title={CARLA: An open urban driving simulator},
  author={Dosovitskiy, Alexey and Ros, German and Codevilla, Felipe and Lopez, Antonio and Koltun, Vladlen},
  booktitle={Conference on Robot Learning},
  pages={1--16},
  year={2017},
  organization={PMLR}
}

@article{hu2025collaborative,
  title={Collaborative perception for connected and autonomous driving: Challenges, possible solutions and opportunities},
  author={Hu, Senkang and Fang, Zhengru and Deng, Yiqin and Chen, Xianhao and Fang, Yuguang},
  journal={IEEE Wireless Communications},
  year={2025},
  publisher={IEEE}
}

@article{ma2026sense4fl,
  author={Ma, Yanan and Hu, Senkang and Fang, Zhengru and Ji, Yun and Deng, Yiqin and Fang, Yuguang},
  journal={IEEE Transactions on Mobile Computing}, 
  title={{Sense4FL}: Vehicular Crowdsensing Enhanced Federated Learning for Object Detection in Autonomous Driving}, 
  year={2026},
  volume={25},
  number={8},
  pages={13004-13018},
  doi={10.1109/TMC.2026.3674333}}

@inproceedings{wang2020train,
  title={Train in germany, test in the usa: Making 3d object detectors generalize},
  author={Wang, Yan and Chen, Xiangyu and You, Yurong and Li, Li Erran and Hariharan, Bharath and Campbell, Mark and Weinberger, Kilian Q and Chao, Wei-Lun},
  booktitle={Proceedings of the IEEE/CVF Conference on Computer Vision and Pattern Recognition},
  pages={11713--11723},
  year={2020}
}

@article{fang2024r,
  author={Fang, Zhengru and Wang, Jingjing and Ma, Yanan and Tao, Yihang and Deng, Yiqin and Chen, Xianhao and Fang, Yuguang},
  journal={IEEE Journal on Selected Areas in Communications}, 
  title={{R-ACP}: Real-Time Adaptive Collaborative Perception Leveraging Robust Task-Oriented Communications}, 
  year={2025},
  volume={},
  number={},
  pages={1-1},
  doi={10.1109/JSAC.2025.3623179}}

@article{fang2024pacp,
  author={Fang, Zhengru and Hu, Senkang and An, Haonan and Zhang, Yuang and Wang, Jingjing and Cao, Hangcheng and Chen, Xianhao and Fang, Yuguang},
  journal={IEEE Transactions on Mobile Computing}, 
  title={{PACP}: Priority-Aware Collaborative Perception for Connected and Autonomous Vehicles}, 
  year={2024},
  volume={23},
  number={12},
  pages={15003-15018},
  doi={10.1109/TMC.2024.3449371}}

@article{chen2024vehicle,
  title={Vehicle as a service ({VaaS}): Leverage vehicles to build service networks and capabilities for smart cities},
  author={Chen, Xianhao and Deng, Yiqin and Ding, Haichuan and Qu, Guanqiao and Zhang, Haixia and Li, Pan and Fang, Yuguang},
  journal={IEEE Communications Surveys \& Tutorials},
  volume={26},
  number={3},
  pages={2048--2081},
  year={2024},
  publisher={IEEE}
}

@inproceedings{ruan2024fully,
  title={Fully test-time adaptation for object detection},
  author={Ruan, Xiaoqian and Tang, Wei},
  booktitle={Proceedings of the IEEE/CVF Conference on Computer Vision and Pattern Recognition},
  pages={1038--1047},
  year={2024}
}

@article{hu2022where2comm,
  title={Where2comm: Communication-efficient collaborative perception via spatial confidence maps},
  author={Hu, Yue and Fang, Shaoheng and Lei, Zixing and Zhong, Yiqi and Chen, Siheng},
  journal={Advances in Neural Information Processing Systems},
  volume={35},
  pages={4874--4886},
  year={2022}
}

@inproceedings{xu2022opv2v,
  title={{OPV2V}: An open benchmark dataset and fusion pipeline for perception with vehicle-to-vehicle communication},
  author={Xu, Runsheng and Xiang, Hao and Xia, Xin and Han, Xu and Li, Jinlong and Ma, Jiaqi},
  booktitle={2022 International Conference on Robotics and Automation (ICRA)},
  pages={2583--2589},
  year={2022},
  organization={IEEE}
}

@incollection{kellerer2004introduction,
  title={Introduction to {NP}-Completeness of knapsack problems},
  author={Kellerer, Hans and Pferschy, Ulrich and Pisinger, David},
  booktitle={Knapsack problems},
  pages={483--493},
  year={2004},
  publisher={Springer}
}

@inproceedings{yoo2025learning,
  title={Learning {3D} Perception from Others' Predictions},
  author={Yoo, Jinsu and Feng, Zhenyang and Pan, Tai-Yu and Sun, Yihong and Phoo, Cheng Perng and Chen, Xiangyu and Campbell, Mark and Weinberger, Kilian and Hariharan, Bharath and Chao, Wei-Lun},
  booktitle={International Conference on Learning Representations},
  volume={2025},
  pages={82610--82630},
  year={2025}
}

@inproceedings{xia2025learning,
  title={Learning to Detect Objects from Multi-Agent LiDAR Scans without Manual Labels},
  author={Xia, Qiming and Lin, Wenkai and Xiang, Haoen and Huang, Xun and Chen, Siheng and Dong, Zhen and Wang, Cheng and Wen, Chenglu},
  booktitle={Proceedings of the Computer Vision and Pattern Recognition Conference},
  pages={1418--1428},
  year={2025}
}

@inproceedings{lang2019pointpillars,
  title={Pointpillars: Fast encoders for object detection from point clouds},
  author={Lang, Alex H and Vora, Sourabh and Caesar, Holger and Zhou, Lubing and Yang, Jiong and Beijbom, Oscar},
  booktitle={Proceedings of the IEEE/CVF Conference on Computer Vision and Pattern Recognition},
  pages={12697--12705},
  year={2019}
}

@inproceedings{wu2024commonsense,
  title={Commonsense prototype for outdoor unsupervised {3D} object detection},
  author={Wu, Hai and Zhao, Shijia and Huang, Xun and Wen, Chenglu and Li, Xin and Wang, Cheng},
  booktitle={Proceedings of the IEEE/CVF Conference on Computer Vision and Pattern Recognition},
  pages={14968--14977},
  year={2024}
}

@article{li2016revisiting,
  title={Revisiting batch normalization for practical domain adaptation},
  author={Li, Yanghao and Wang, Naiyan and Shi, Jianping and Liu, Jiaying and Hou, Xiaodi},
  journal={arXiv preprint arXiv:1603.04779},
  year={2016}
}

@inproceedings{liu2020who2com,
  title={Who2com: Collaborative perception via learnable handshake communication},
  author={Liu, Yen-Cheng and Tian, Junjiao and Ma, Chih-Yao and Glaser, Nathan and Kuo, Chia-Wen and Kira, Zsolt},
  booktitle={2020 IEEE International Conference on Robotics and Automation (ICRA)},
  pages={6876--6883},
  year={2020},
  organization={IEEE}
}

@inproceedings{wang2020v2vnet, 
title={{V2VNet}: Vehicle-to-vehicle communication for joint perception and prediction}, 
author={Wang, Tsun-Hsuan and Manivasagam, Sivabalan and Liang, Ming and Yang, Bin and Zeng, Wenyuan and Urtasun, Raquel}, 
booktitle={European Conference on Computer Vision}, 
pages={605--621}, 
year={2020}, 
organization={Springer} 
}

@inproceedings{roychowdhury2019automatic,
  title={Automatic adaptation of object detectors to new domains using self-training},
  author={RoyChowdhury, Aruni and Chakrabarty, Prithvijit and Singh, Ashish and Jin, SouYoung and Jiang, Huaizu and Cao, Liangliang and Learned-Miller, Erik},
  booktitle={Proceedings of the IEEE/CVF Conference on Computer Vision and Pattern Recognition},
  pages={780--790},
  year={2019}
}

@INPROCEEDINGS{tao2024directed,
  author={Tao, Yihang and Hu, Senkang and Fang, Zhengru and Fang, Yuguang},
  booktitle={2025 IEEE International Conference on Robotics and Automation (ICRA)}, 
  title={{Directed-CP}: Directed Collaborative Perception for Connected and Autonomous Vehicles via Proactive Attention}, 
  year={2025},
  volume={},
  number={},
  pages={7004-7010},
  doi={10.1109/ICRA55743.2025.11127818}}

@inproceedings{boudiaf2022parameter,
  title={Parameter-free online test-time adaptation},
  author={Boudiaf, Malik and Mueller, Romain and Ben Ayed, Ismail and Bertinetto, Luca},
  booktitle={Proceedings of the IEEE/CVF Conference on Computer Vision and Pattern Recognition},
  pages={8344--8353},
  year={2022}
}

@article{zhang2022memo,
  title={{MEMO}: Test time robustness via adaptation and augmentation},
  author={Zhang, Marvin and Levine, Sergey and Finn, Chelsea},
  journal={Advances in Neural Information Processing Systems},
  volume={35},
  pages={38629--38642},
  year={2022}
}

@article{you2022unsupervised,
  title={Unsupervised adaptation from repeated traversals for autonomous driving},
  author={You, Yurong and Phoo, Cheng Perng and Luo, Katie and Zhang, Travis and Chao, Wei-Lun and Hariharan, Bharath and Campbell, Mark and Weinberger, Kilian Q},
  journal={Advances in Neural Information Processing Systems},
  volume={35},
  pages={27716--27729},
  year={2022}
}

@article{yang2023how2comm,
  title={How2comm: Communication-efficient and collaboration-pragmatic multi-agent perception},
  author={Yang, Dingkang and Yang, Kun and Wang, Yuzheng and Liu, Jing and Xu, Zhi and Yin, Rongbin and Zhai, Peng and Zhang, Lihua},
  journal={Advances in Neural Information Processing Systems},
  volume={36},
  pages={25151--25164},
  year={2023}
}

@inproceedings{gao2025stamp,
  title={{STAMP}: Scalable task-and model-agnostic collaborative perception},
  author={Gao, Xiangbo and Xu, Runsheng and Li, Jiachen and Wang, Ziran and Fan, Zhiwen and Tu, Zhengzhong},
  booktitle={International Conference on Learning Representations},
  volume={2025},
  pages={54656--54676},
  year={2025}
}

@article{cao2024exploring,
  title={Exploring test-time adaptation for object detection in continually changing environments},
  author={Cao, Shilei and Zheng, Juepeng and Liu, Yan and Zhao, Baoquan and Yuan, Ziqi and Li, Weijia and Dong, Runmin and Fu, Haohuan},
  journal={arXiv preprint arXiv:2406.16439},
  year={2024}
}

@inproceedings{ince2021semi,
  title={Semi-automatic annotation for visual object tracking},
  author={Ince, Kutalmis Gokalp and Koksal, Aybora and Fazla, Arda and Alatan, A Aydin},
  booktitle={Proceedings of the IEEE/CVF International Conference on Computer Vision},
  pages={1233--1239},
  year={2021}
}

@inproceedings{bai2025annotation,
  title={Annotation Methods for Object Detection: A Comparative Analysis from Manual Labeling to Automated Annotation Technologies},
  author={Bai, Dongmei and Wei, Shijia and He, Xu and Yu, Qiaoming},
  booktitle={2025 5th International Conference on Artificial Intelligence and Industrial Technology Applications (AIITA)},
  pages={1473--1479},
  year={2025},
  organization={IEEE}
}

@inproceedings{reza2025segbuilder,
  title={Segbuilder: A semi-automatic annotation tool for segmentation},
  author={Reza, Md Alimoor and Manley, Eric and Chen, Sean and Chaudhary, Sameer and Elafros, Jacob},
  booktitle={2025 IEEE/CVF Winter Conference on Applications of Computer Vision (WACV)},
  pages={8494--8503},
  year={2025},
  organization={IEEE}
}

@inproceedings{hu2024adaptive,
  title={Adaptive communications in collaborative perception with domain alignment for autonomous driving},
  author={Hu, Senkang and Fang, Zhengru and An, Haonan and Xu, Guowen and Zhou, Yuan and Chen, Xianhao and Fang, Yuguang},
  booktitle={GLOBECOM 2024-2024 IEEE Global Communications Conference},
  pages={746--751},
  year={2024},
  organization={IEEE}
}

@inproceedings{bengio2009curriculum,
  title={Curriculum learning},
  author={Bengio, Yoshua and Louradour, J{\'e}r{\^o}me and Collobert, Ronan and Weston, Jason},
  booktitle={Proceedings of the 26th Annual International Conference on Machine Learning},
  pages={41--48},
  year={2009}
}

@inproceedings{zhu2023curricular,
  title={Curricular object manipulation in lidar-based object detection},
  author={Zhu, Ziyue and Meng, Qiang and Wang, Xiao and Wang, Ke and Yan, Liujiang and Yang, Jian},
  booktitle={Proceedings of the IEEE/CVF Conference on Computer Vision and Pattern Recognition},
  pages={1125--1135},
  year={2023}
}

@article{ma2026birdcast,
  title={Birdcast: Interest-aware {BEV} Multicasting for Infrastructure-assisted Collaborative Perception},
  author={Ma, Yanan and Fang, Zhengru and Tao, Yihang and Guo, Yu and Deng, Yiqin and Chen, Xianhao and Fang, Yuguang},
  journal={arXiv preprint arXiv:2604.00701},
  year={2026}
}

@article{ma2026update,
  title={Update the Unseen Only: Minimizing {AoI} for Collaborative Perception through Online Learning},
  author={Ma, Yanan and Zhao, Zhuoyi and Fang, Zhengru and An, Haonan and Chen, Xianhao and Fang, Yuguang},
  journal={arXiv preprint arXiv:2607.20967},
  year={2026}
}

\end{document}